\documentclass[runningheads]{llncs}
\usepackage[T1]{fontenc}
\usepackage{multirow}
\usepackage{makecell}
\usepackage{tikz}
\usepackage{mathrsfs}
\usetikzlibrary{positioning}

\usepackage{amsmath}
\usepackage{amsfonts}
\usepackage{float}
\usepackage[utf8]{inputenc}
\usepackage{graphicx}
\usepackage{caption}
\usepackage{subcaption}
\usepackage{diagbox}
\usepackage{anyfontsize}
\usepackage{booktabs}
\usepackage{graphicx}
\usepackage{url}

\begin{document}
\title{Know Yourself Better: Diverse Object-Related Features Improve Open
Set Recognition}
\titlerunning{Know Yourself Better}
%
\author{Jiawen Xu\inst{1} \and Margret Keuper\inst{2} \and Odej Kao\inst{1}}
%
%
\institute{Technical University Berlin, Einsteinufer 17,
10587 Berlin, Germany \\
\email{jiawen.xu@campus.tu-berlin.de}
\and
University of Mannheim, B6 26, 68159 Mannheim, Germany \\
\email{keuper@uni-mannheim.de}
}
\maketitle              
\begin{abstract}
Conventional neural classifiers are trained under the close set assumption and therefore struggle to identify samples from previously unseen categories, often producing overconfident predictions in realistic deployment scenarios. Open set recognition (OSR) addresses this challenge by enabling models to detect novel classes at inference time. While numerous methods have been proposed and shown promising empirical performance, the representation-learning mechanisms underlying successful OSR remain insufficiently understood.
In this work, we investigate OSR from the perspective of feature diversity. Through a series of controlled experiments, we demonstrate that representations encoding a broader range of object-related features consistently lead to improved OSR performance. We further analyze supervised contrastive learning (SupCon) and show that its temperature parameter significantly influences the representation learning process by assigning different emphasis to sample pairs of varying similarities. As a result, models trained with different temperatures learn complementary representations of the data.
Motivated by these findings, we propose a temperature-based ensemble framework that aggregates the outputs of SupCon models trained with different temperatures. Extensive experiments on standard OSR benchmarks demonstrate that the proposed approach consistently outperforms strong baselines.

\end{abstract}

\section{Introduction} \label{sec-intro}
Deep neural networks have achieved remarkable success across a wide range of classification tasks, including video classification~\cite{karpathy2014large}, sentiment analysis~\cite{zhang2018deep}, and fault diagnosis~\cite{lei2020applications}. However, in real-world applications, the set of categories encountered during deployment is often difficult to enumerate exhaustively and may evolve over time, resulting in a mismatch between the training and test distributions. Samples belonging to previously unseen categories, commonly referred to as open set classes, may therefore be misclassified as one of the known classes, since conventional classifiers lack the ability to explicitly express uncertainty through an ``unknown'' prediction.
This challenge motivates the study of open-set recognition (OSR), which aims to determine whether a test sample belongs to one of the known classes observed during training or originates from an unseen class. Owing to its importance in safety-critical and open world applications, OSR has attracted considerable attention and remains an active area of research~\cite{asg_Yu17,GOpenMax_Ge17,neal2018open,hassen2020learning,dhamija2018reducing,yang2020convolutional,vaze2022openset,wang2024exploring}.
In this work, we investigate OSR from the perspective of feature diversity, an aspect that has received relatively limited attention in the existing literature. Through controlled experiments, we study the relationship between the diversity of features encoded in learned representations and OSR performance, and leverage the resulting insights to develop an effective OSR approach.
With the rapid adoption of foundation models (FMs) in recent years, one may question whether open set recognition remains an important research problem. While FMs often exhibit strong performance on a broad range of unseen concepts, this capability is largely attributable to their training on web-scale datasets rather than an inherent ability to reliably quantify uncertainty or identify all outlier samples. Consequently, the principles investigated in this work are a relevant and complementary research direction compared to the use of FMs for outlier detection~\cite{li2024learning,ming2022delving}. In particular, our analysis focuses on how the representation learning can be shaped to capture more diverse features, a property that may benefit a wide range of tasks beyond OSR~\cite{kundu2025boosting}.
Unless otherwise specified, in the following text, we refer to samples from known training classes as \emph{inliers} and samples from unseen classes as \emph{outliers} throughout the paper.

\begin{figure}
    \centering
    \includegraphics[width=0.5\textwidth]{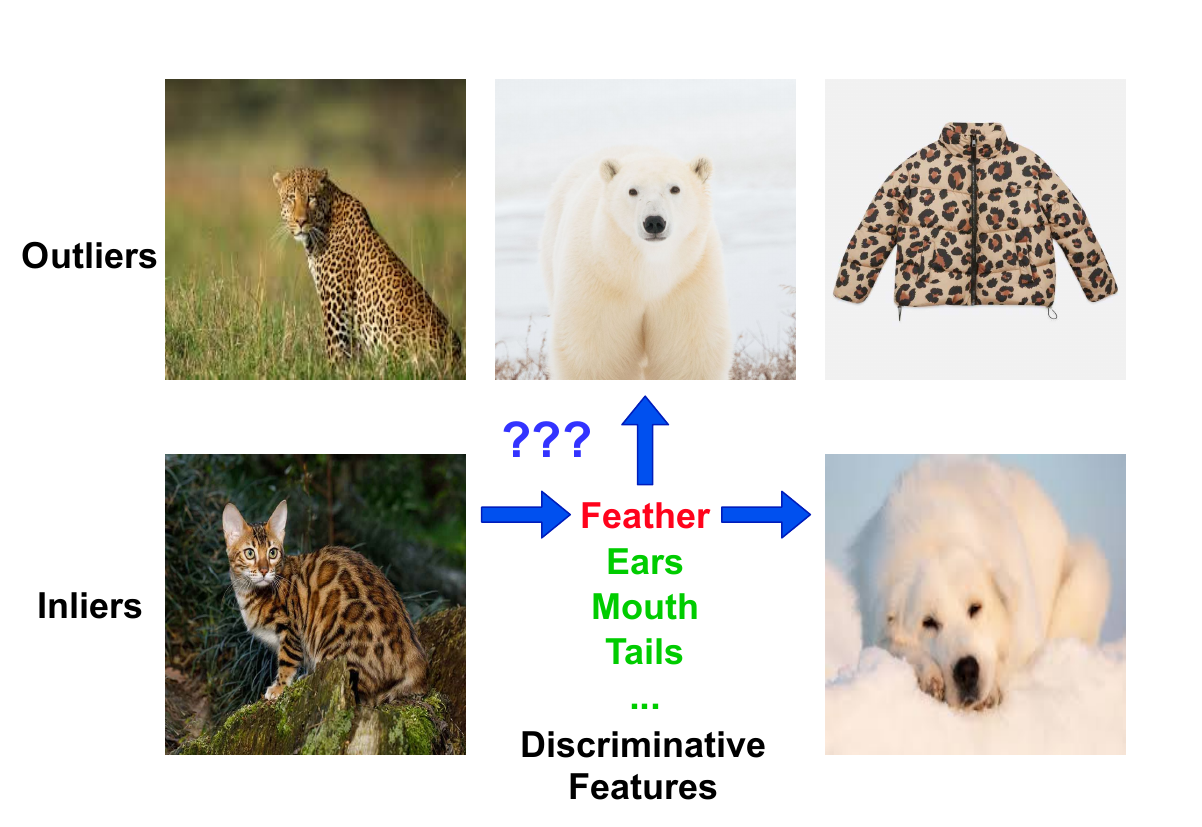}
    \caption{Illustration of the intrinsic link between feature diversity and OSR performance. Consider the following scenario: Inliers, cats and dogs, in this example, can be accurately classified by leveraging a discriminative feature such as feather patterns. However, when faced with outliers, leopards, polar bears, and leopard-patterned jackets, relying solely on the feather textures becomes problematic, especially when these outliers exhibit high similarity in these particular feather textures. In such a case, additional features, such as the shapes of the ears and tails, have to be learned to enable the model to discern and handle a wider range of outliers effectively.}
    \label{fig-intutive}
\end{figure}

Constructing a training dataset that comprehensively covers all possible outlier classes is inherently infeasible, as the open-set space is effectively unbounded and may contain categories that are inaccessible during training. To address this challenge, several OSR approaches employ generative models to synthesize pseudo-outliers~\cite{asg_Yu17,GOpenMax_Ge17,neal2018open}. These methods augment the training set with generated outlier samples alongside the inlier data.
The synthesized outliers are typically designed to be challenging examples that closely resemble the inlier classes. By introducing such hard-to-distinguish samples during training, the model is encouraged to learn more diverse discriminative features, thereby improving its ability to distinguish inliers from these synthesized outliers.
In the study of~\cite{vaze2022openset}, it has been observed that data augmentation and label smoothing can improve OSR. However, the specific reasons behind these improvements remain unclear.
Among these strategies, data augmentation can not only provide more training samples, but also facilitates the capture of informative yet challenging-to-learn features~\cite{shen2022data}. These features are often susceptible to neglect due to their relatively minor contributions for reducing loss values during training with the original datasets alone. 
Label smoothing alleviates the excessive concentration of representations around class prototypes and reduces the pressure toward class collapse, thereby requiring more diverse features to be encoded in the representations.

Building upon the above observations, we hypothesize that increasing the diversity of \emph{\textbf{object-related features}} encoded in the representations can improve open set recognition performance. 
We define feature diversity as the extent to which a representation captures multiple semantically meaningful object attributes that can contribute to distinguishing samples, rather than relying on a limited subset of discriminative cues.
Figure~\ref{fig-intutive} provides an intuitive illustration of how richer feature representations can facilitate the detection of outliers. To further investigate this hypothesis, we conduct a series of controlled experiments in Section~\ref{sec-toy-example}, demonstrating that capturing a broader range of object-related features is closely associated with improving OSR performance.

Previous studies have shown that the temperature parameter in self-supervised contrastive learning influences the optimization dynamics by assigning different weights to sample pairs with varying similarity levels~\cite{wang2021understanding}. In this work, we extend this perspective to supervised contrastive learning (SupCon) and analyze how the temperature affects the relative emphasis placed on intra-class and inter-class pairs with different similarities. Our analysis suggests that different temperature values induce distinct representation learning behaviors, leading the models to capture complementary characteristics of the data.

Based on the above findings, we propose a temperature-based ensemble framework that leverages supervised contrastive learning models trained with different temperature values to obtain more diverse representations for open set recognition, as detailed in Section~\ref{sec-method}. Experimental results on standard OSR benchmarks demonstrate that, despite its simplicity, the proposed approach achieves competitive or superior performance compared with widely used baselines.

Our contributions are summarized as follows:

\begin{itemize}
\item We conduct a series of controlled experiments that reveal a positive correlation between the diversity of object-related features encoded in learned representations and open-set recognition performance.

\item We provide an analysis of the temperature parameter in supervised contrastive learning and show that different temperatures induce distinct representation learning behaviors, leading to complementary feature representations.

\item Motivated by these findings, we propose a simple temperature-based ensemble framework for OSR that exploits the complementary representations learned under different temperatures. Extensive experiments on standard benchmarks demonstrate the effectiveness of the proposed approach.

\end{itemize}

Our hypothesis and findings regarding representation learning for open-set recognition are most closely related to those of~\cite{dietterich2022familiarity} and~\cite{wang2024exploring}. While these works provide valuable theoretical and conceptual insights, our study investigates the role of feature diversity in OSR from a complementary representation-learning perspective.
Specifically,~\cite{dietterich2022familiarity} proposes the familiarity hypothesis, which suggests that successful OSR methods operate by detecting the absence of features learned from the inlier classes. Meanwhile,~\cite{wang2024exploring} shows that, under the empirical risk minimization framework for classifiers trained with cross-entropy loss, reducing the conditional entropy of model predictions given the learned representations can lower open set risk.
In contrast, our work focuses on the relationship between feature diversity and OSR performance through controlled empirical studies. Rather than relying on a particular training objective or outlier detection strategy, our analysis is conducted from the perspective of representation learning and is largely independent of specific classification paradigms or OSR methods.

\section{Related Work}
\label{sec-background}

\subsection{Open Set Recognition}
Open set recognition aims to identify samples from novel classes during machine learning classifier inference 
In recent years, there has been a surge in the development of OSR methods tailored for deep neural networks. Bendale et. al have firstly revealed that the activation patterns in the penultimate layers of the deep classifiers (features) can exhibit distinct characteristics for inliers and outliers~\cite{bendale2016towards}.  The features of the inliers are then modeled with the \emph{Weibull} distribution to reject outliers.
Subsequently, numerous OSR methods follow this paradigm, which detects outliers using the deep features output by trained classifiers \cite{hassen2020learning,dhamija2018reducing,yang2020convolutional,perera2020generative,millerclass,chen_2020_ECCV,kodama2021open,xu2023contrastive}. To better model inliers and possibly outliers using deep features, novel learning strategies have been proposed. For example, in~\cite{hassen2020learning},~\cite{yang2020convolutional},~\cite{chen_2020_ECCV}, and~\cite{millerclass}, novel learning objectives are designed with the principle of enhancing intra-class tightness and inter-class separation in feature space. 
Outliers sampled from extra datasets are applied to train the model in~\cite{dhamija2018reducing}. Besides, some works synthesizes outliers using, for example, generative models, to train the classifiers altogether with the inliers~\cite{asg_Yu17,GOpenMax_Ge17,neal2018open}.
In addition to the above paradigm that relies on the discriminative classifiers, some works apply, for example, generative models to learn the representations of the inliers~\cite{oza2019c2ae,yoshihashi2019classification,zhang2020hybrid,cao2021open}. In~\cite{oza2019c2ae}, a class-conditioned auto-encoder is trained on inliers, and their reconstruction errors are modeled using extreme value distributions to identify outliers during inference. 

\subsection{Contrastive Representation Learning}

Contrastive learning is popular for its generalizable representation learning ability and was first proposed for self-supervised learning~\cite{lee2018simple}, which is known as self-supervised contrastive learning (SSL). The main idea behind contrastive learning is to enclose the representations between the different views of the same data sample.
There are massive downstream applications, such as semi-supervised learning~\cite{chen2020big}, continual learning~\cite{fini2022self}, and few-shot learning~\cite{an2021conditional}.
Contrastive learning has been extended to supervised fashion in~\cite{khosla2020supervised}, which is named supervised contrastive learning (SupCon) will be introduced in details in Section~\ref{sec-method-pre}.
SupCon is reported to be able to learn more generalizable discriminative representations than its cross-entropy counterpart. 
Specifically, SupCon has been applied to open set recognition in~\cite{kodama2021open,xu2023contrastive}. In~\cite{xu2023contrastive}, the mixup strategy~\cite{zhang2017mixup} is applied to augment the positive pairs. We include~\cite{kodama2021open} and~\cite{xu2023contrastive} in the baselines of our method.

\subsection{Representation Aggregation} \label{subsec-ensembling}

Representation aggregation has been applied to various model generalization problems in deep learning, such as few-shot learning~\cite{han20193d2seqviews}, adversarial robustness~\cite{fort2024ensemble}, and open set recognition~\cite{wang2024exploring} or out-of-distribution detection. These works often aggregate the representations with high variance. For example, in~\cite{lee2018simple}, representations from layers of different depths in neural networks are aggregated for out-of-distribution detection. Under such an intuition that the feature diversity can be enriched.
Representation aggregation is an intuitive and natural way to increase feature diversity 
and can be applied to many other domains in neural networks, such as model architecture design, like \emph{ResNet}~\cite{he2016deep}. 

\section{Study} \label{sec-toy-example}

To investigate whether learning more diverse features of the inlier objects can improve open set recognition, a group of controlled experiments is conducted. We train and test multilayer convolutional neural networks (CNNs) that classify synthetic images as shown in Figure~\ref{fig-three-toy}. The network architecture is detailed in Appendix~\ref{app-toy-archicture}. The synthetic images are with a resolution of 64x64 and with the content of circles and rectangles. The radius of the circles is sampled from a uniform distribution with an interval from 10 to 30. Similarly, the hight and width of the rectangles respectively follow uniform distribution, both with the interval from 10 to 30.
The centers of the circles and rectangles are randomly and uniformly selected within the image scope. 
Two experiments (E1 and E2) are conducted under the conditions outlined in Table~\ref{tab-toy-example}.
In E1, the inlier classes consist of blue circles and red rectangles, while the outlier class comprises blue rectangles. To eliminate the background influence, the background of all images in E1 and E2 is set to black. In such a setting, the model can perfectly distinguish between the two inlier classes using colors only, and it is known that CNNs exhibit a bias towards color over shape~\cite{geirhos2018imagenet,singh2020assessing}. Thus, we can safely assume that the model in E1 relies primarily on the color features.
In E2, a third inlier class, the red circle, is introduced. In such a case, accurately classifying all inlier classes requires the model to incorporate shape information. Therefore, models in E2 are supposed to be expected to learn more features that include both color and shape. For both E1 and E2, there are 100 images of each class for training, and 50 images of each for testing. After 30 epochs of training, the inlier classification accuracy in E1 and E2 is $100\%$ and $95.33\%$, respectively, as listed in Table~\ref{tab-toy-example}.

\begin{figure}[htbp]
\centering
\begin{minipage}[c]{0.6\textwidth}
\centering
    \begin{tabular}{ cccc } 
    \toprule[1.5pt]
    Index & Inliers &  Outliers &  \makecell{Inlier \\ Accuracy}    \\
    \midrule[1pt]
    E1  & \makecell{\textcolor{blue}{Blue Circles}, \\ \textcolor{red}{Red Rectangles}} &\makecell{\textcolor{blue}{Blue} \\ \textcolor{blue}{Rectangles}}  & 100\%   \\[5mm]  
    
    E2  & \makecell{\textcolor{blue}{Blue Circles}, \\ \textcolor{red}{Red Rectangles} \\ \textcolor{red}{Red Circles}} & \makecell{\textcolor{blue}{Blue} \\ \textcolor{blue}{Rectangles}}   &  95.33\%            \\[5mm] 
    \bottomrule[1.5pt]
\end{tabular}
\vspace{6pt}
\captionof{table}{Settings for the two groups of controlled experiments in Section \ref{sec-toy-example}. Blue circles and red rectangles are inliers in E1 and red circles are additionally introduced in E2. The inlier classification accuracy for E1 and E2 is $100\%$ and $95.33\%$, respectively. And the outliers are blue rectangles for both E1 and E2.}
\label{tab-toy-example}
\end{minipage}
\hfill
\begin{minipage}[c]{0.38\textwidth}
\centering
\fbox{\includegraphics[width=0.3\textwidth]{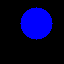}}
\fbox{\includegraphics[width=0.3\textwidth]{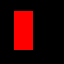}} \\[0.5em]
\fbox{\includegraphics[width=0.3\textwidth]{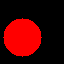}}
\fbox{\includegraphics[width=0.3\textwidth]{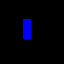}}
\vspace{6pt}
\captionof{figure}{Examples from the synthetic dataset in the controlled experiments, which are (from left to right, up to down) blue circle, red rectangle, red circle, and blue rectangle. All backgrounds are set to be black.}
\label{fig-three-toy}
\end{minipage}
\end{figure}

\begin{figure}[htbp]
\centering
\begin{minipage}[c]{0.45\textwidth}
\centering
\raisebox{-0.5\height}{
\fbox{\includegraphics[width=0.35\textwidth]{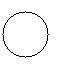}} 
\fbox{\includegraphics[width=0.35\textwidth]{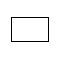}}} 
\vspace{6pt}
\captionof{figure}{Examples from the synthetic dataset in the controlled experiments. The circles and rectangles are not filled to evaluate if the model can recognize shapes.}
\label{fig-toy-shape}
\end{minipage}
\hfill
\begin{minipage}[c]{0.5\textwidth}
\centering
    \begin{tabular}{cccc} 
    \toprule[1.5pt]
     & conv1 &   linear1 &  linear2    \\
    \midrule[1pt]
    E1  & $72.75\%$ &  $64\%$ &  $62\%$ \\
    E2  & $\mathbf{83.33\%}$ & $\mathbf{76\%}$ & $\mathbf{72\%}$   \\
    \bottomrule[1.5pt]
\end{tabular}
\vspace{6pt}
\captionof{table}{Testing accuracy of the models in E1 and E2 after finetuning on shapes only synthetic data. The first row indicates till which layers the parameters are frozen. Bold indicates the better results under each layer frozen setting.}
\label{tab-shape}
\end{minipage}
\end{figure}

To verify that it is harder for the model in E1 to learn shape features, we finetune the trained models in E1 and E2 by freezing the early layers using the synthetic images without color filling, as demonstrated in Figure~\ref{fig-toy-shape}. The finetuned models can hence no longer rely on the colors to classify the data. The results are shown in Table~\ref {tab-shape}. There are three groups of layer frozen settings, namely \emph{conv1}, \emph{linear1}, and \emph{linear2} (the layer names are in Table~\ref{tab-toy-architecture} in Appendix~\ref{app-toy-archicture}), which means that the models are frozen till layer \emph{conv1}, \emph{linear1}, and \emph{linear2}. The results show that the models in E2 can always achieve higher accuracy, indicating better shape features encoded in their frozen layers. These results align with our assumptions above. In addition, we can draw two extra conclusions, which are, however, beyond the scope of this work. First, the shape features are better encoded in early layers for both E1 and E2, which aligns with the Information Bottleneck principle~\cite{tishby2015deep}. Second, even in E2, the finetuning accuracy is not $100\%$, which can indicate that the model is biased towards color. 
And we think it is the reason why the inlier testing accuracy in E2 is lower than in E1.
To further verify this, we directly test the models without finetuning on the data in Table~\ref {tab-shape}, and the accuracy is $50\%$ for both E1 and E2. We believe the reason lies in that the most weights connecting layer \emph{linear2} and \emph{linear3} are biased towards color, and the black-white images in Table~\ref {tab-shape} hence show very few differences even for the E2 model. 
We leave these two extra findings for future work.

We then test the open set recognition performances on the E1 and E2 models. We apply three detection methods, which are maximum softmax probability~\cite{bendale2016towards}, \emph{Mahalanobis} distances~\cite{liu2020few,lee2018simple}, and feature norm~\cite{vaze2022openset}, on top of the trained models to detect outliers. 
These three methods are abbreviated as \emph{MSP}, \emph{M-distance}, and \emph{Norm} in the following text.
In the MSP method, the maximum softmax probability output by the model is the score for recognizing outliers. The higher the score is, the more confidence the model has on its prediction, and the less likely the testing sample outlier is. 
In the \emph{M-distance} method, we record the \emph{Mahalanobis} distances, $M_{c}$, between the testing sample representations with their closest class centers in the training set to measure their similarities. 
The lower the similarity is, the lower the probability that the testing sample belong to any class of the training data, and hence it is more likely to be an outlier.
For the \emph{norm} method, the $L_2$ norms of the representations are computed as the OSR score. The norms of the inliers are supposed to be higher than those of the outliers~\cite{vaze2022openset}.
We take the representations from the layer \emph{linear2} as inputs to the \emph{M-distance} and \emph{norm} methods. 
We apply the Area under the ROC curve (AUROC) as metric to measure the OSR performances (refer Section~\ref{subsec-exp-settings} for an introduction). We take blue rectangles as outliers, which require the model to know shapes for a successful open set recognition. There are 50 outlier samples in the outlier testing set, along with 50 samples in each inlier testing set.

\begin{figure}[htbp]
\centering 
\begin{minipage}[c]{0.45\textwidth} 
\centering
\begin{tabular}{cccc} 
    \toprule[1.5pt]
    AUROC & MSP & M-distance & Norm \\
    \midrule[1pt]
    E1 & $98.2\%$ & $89.7\%$ & $81.1\%$ \\
    E2 & $\mathbf{99.1\%}$ & $\mathbf{91.5\%}$ & $\mathbf{97\%}$ \\
    \bottomrule[1.5pt]
\end{tabular}
\vspace{6pt}
\captionof{table}{AUROC for OSR detection using predicted probabilities or the outputs of the \emph{linear2} layer as input representations. The model in E2 can show superior performances when with all detection methods, MSP, Mahalanobis distance, and feature norm.}
\label{table-toy-osr}
\end{minipage} 
\hfill
\begin{minipage}[c]{0.53\textwidth} 
\centering
    \includegraphics[width=\linewidth]{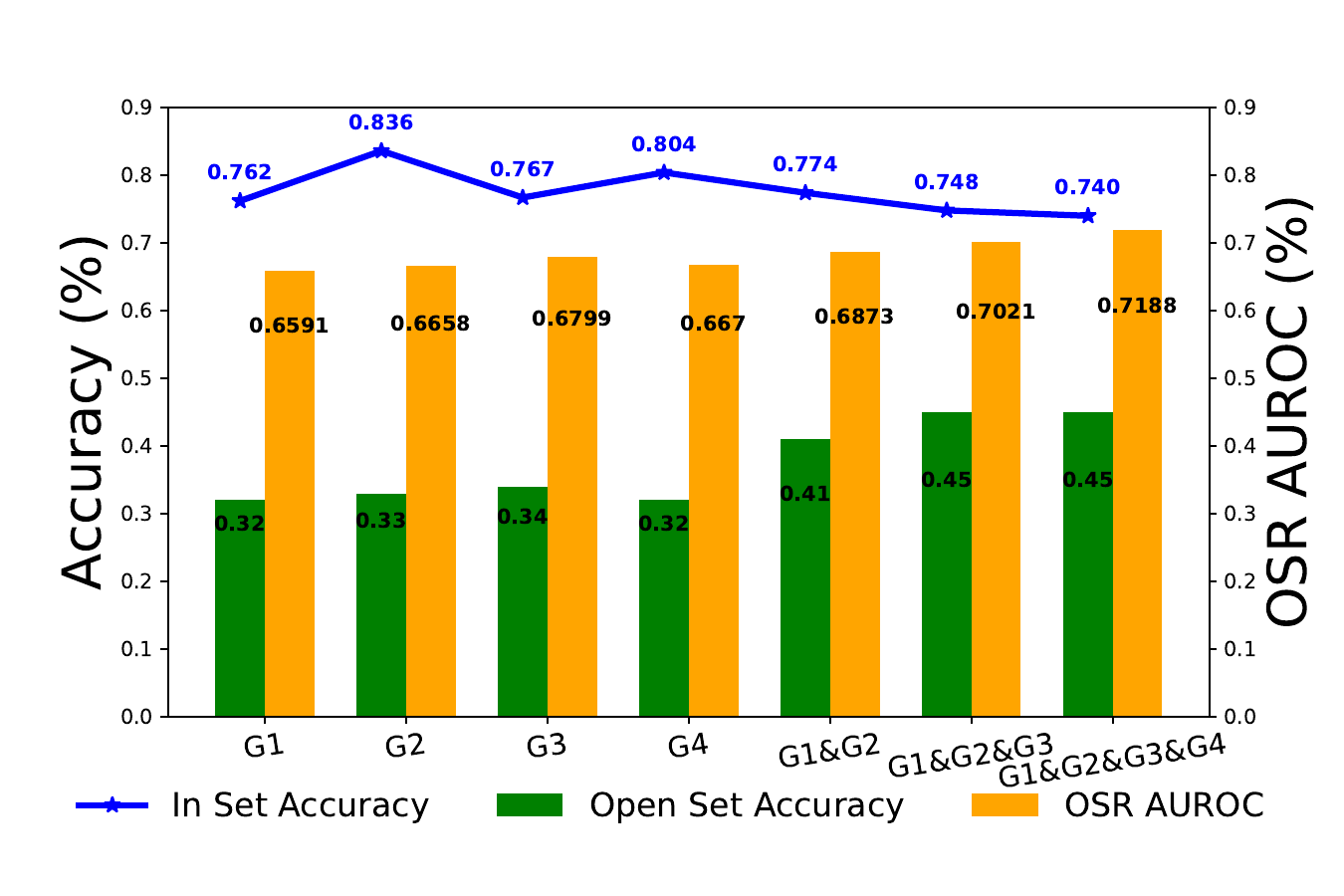} 
    \caption{Open set accuracy and OSR AUROC of the models trained with differet groups of data in CIFAR100. The more data that the models trained more, the more diverse features they learned and the better they are at detecting open sets. }
    \label{fig-cifar100}
\end{minipage}
\end{figure}

The OSR results are shown in Table~\ref {table-toy-osr}. The AUROC in E2 is higher than E1 with all three detection methods.
Since the E2 model can classify red rectangles and red circles, it can then be safely believed to discriminate blue circles and blue rectangles well. 
However, the AUROC values of the E1 model are not $50\%$, which means the E1 model encodes some shape features, which are for sure less effective than those in E2. It aligns with the results in Table~\ref{tab-shape} that the finetuned model on top of the E1 linear2 outputs can still distinguish shapes to some distance ($62\%$).
Moreover, the results in Table~\ref {table-toy-osr} vary greatly between the detection methods. We think MSP and norm methods are more tolerant to the noise and redundancy in the representations. 

Furthermore, to support the generality of the above observations, we conduct additional experiments on the more challenging CIFAR-100 dataset, which consists of 100 semantic classes belonging to 20 macro-classes, with four classes for each. 
Unlike the toy dataset considered previously, where the learned semantic features can be explicitly controlled, it is impractical to directly manipulate the models to learn specific human-interpretable features on CIFAR-100. Instead, we encourage the models to learn feature representations of varying diversity by controlling the number of classes included during training.
Specifically, we partition CIFAR-100 into five groups (G1--G5), each containing one class from every macro-class. The detailed split is provided in Table~\ref{tab-cifar100-group} in Appendix~\ref{app-cifar100}. Based on these groups, we train seven classifiers using cross-entropy loss on G1, G2, G3, G4, G1\&G2, G1\&G2\&G3, and G1\&G2\&G3\&G4, respectively.
Previous work has shown that increasing the number of training classes can improve the generalization and reliability of learned representations~\cite{zhu2022learning}. In our setting, as additional groups are incorporated into the training data while maintaining comparable inlier classification performance, the models are required not only to distinguish between coarse-grained macro-classes but also to discriminate among increasingly fine-grained classes within each macro-class. This increased discrimination difficulty encourages the extraction of more diverse features.

We treat G5 as outlier samples and evaluate the open-set recognition (OSR) performance of the seven classifiers. To further assess the diversity of the learned representations, we also measure the classification accuracy of each model on the outlier classes. The results are presented in Figure~\ref{fig-cifar100}. As the number of training classes increases, the models consistently achieve higher accuracy on the outlier classes and improved OSR performance, while maintaining comparable inlier classification accuracy. These findings support our hypothesis that learning more diverse features leads to better open-set recognition performance.

In summary, the above experiments support our hypothesis that encoding more discriminative and diverse features in the learned representations can improve OSR performance. Moreover, the results reveal several factors that influence this objective.
First, the models tend to exhibit biases toward certain feature types. In our experiments, classifiers often rely heavily on color cues; more generally, deep neural networks are known to be biased toward texture rather than shape information~\cite{geirhos2020shortcut}, particularly when such cues are sufficient for accurate inlier classification. Nevertheless, the neglected information may still be preserved in intermediate or early-layer representations and can potentially be exploited for OSR.
Second, increasing the number of training classes encourages the model to learn more diverse features. As a result, the learned representations lead to better OSR performance and exhibit improved generalization ability.
Consequently, developing methods that encourage feature diversity in the learned representations is a promising direction for improving OSR.

\section{Method} \label{sec-method}

Based on the findings above, we propose a feature ensembling method based on supervised contrastive learning for open-set recognition. We first analyze the representation learning behavior of SupCon under different temperature hyperparameters. Building on these observations, we then introduce our method, which ensembles representations extracted from models trained with varying temperatures.

\subsection{Supervised Contrastive Learning} \label{sec-method-pre}

\subsubsection{General Learning Objective} \label{subsubsec-overall}
Supervised contrastive learning (SupCon)~\cite{khosla2020supervised} extends self-supervised contrastive learning~\cite{chen2020simple} to the supervised setting. In SupCon, positive pairs are formed not only by the original and augmented views of an anchor sample, but also by samples from the same class. In contrast, samples belonging to different classes are treated as negative pairs. The corresponding contrastive loss is defined in Equation~\eqref{equ-contra-loss}:

\begin{equation}
\begin{split}
    & \mathcal{L}_\mathit{SupCon}  = \sum_{i \in I} \frac{-1}{|P(i)|} \sum\limits_{p \in P(i)} \log \frac{\exp \left( \textbf{z}_i \cdot \textbf{z}_{p} / \tau \right)}{\sum\limits_{a \in A(i)} \exp \left( \textbf{z}_i \cdot \textbf{z}_a / \tau \right)}  
    \label{equ-contra-loss}
\end{split}
\end{equation}

In Equation~\eqref{equ-contra-loss}, $\mathbf{z}_i$ denotes the representation of the anchor sample indexed by $i$. The set $P(i)$ contains the indices of all positive samples associated with the anchor, represented by $\mathbf{z}_p$, while $N(i)$ denotes the set of negative samples. The set $A(i)$ includes all samples except the anchor itself, i.e., $A(i) = P(i) \cup N(i)$. The temperature parameter rescales pairwise similarities and consequently alters the distribution of gradients over positive and negative pairs, thereby changing the optimization emphasis on hard and easy samples.

\subsubsection{Temperature and Representation Learning of SupCon} \label{subsubsec-temp-feature}

The role of the temperature parameter $\tau$ in contrastive learning has been extensively studied in the context of self-supervised learning~\cite{robinson2021can,kukleva2022temperature,wang2021understanding}. In this work, we investigate its effects in the supervised contrastive learning (SupCon) setting.

To understand how $\tau$ influences the feature learning process, we analyze the gradients of ${L}_\mathit{SupCon}$ with respect to the similarities of positive and negative pairs, denoted by $s_{ip}$ and $s_{in}$, respectively, as shown in Equation~\eqref{equ-gradient-sip} and~\eqref{equ-gradient-sin}. Here, $s_{ip}$ and $s_{in}$ represent the cosine similarities between the anchor representation and its positive and negative pairs, respectively, i.e., $s_\mathit{ip} = \mathbf{z}_i \cdot \mathbf{z}_p$ and $s\mathit{in} = \mathbf{z}_i \cdot \mathbf{z}_n$. Moreover, $P_{i,j}$ denotes the \emph{Softmax} probability associated with sample $j$ under temperature scaling, defined as $P_{i,j} = \frac{\exp(\frac{\mathbf{z}_{i,j}}{\tau})}{\sum\limits_{a \in A{i}}\exp(\frac{\mathbf{z}_{i,a}}{\tau})}$.
The detailed gradient derivation of Equation~\eqref{equ-gradient-sip} and~\eqref{equ-gradient-sin} is in Appendix \ref{app-gradients}.

\begin{equation}
        g_{ip}
        =\frac{\partial  L_{SupCon}^{i}}{\partial s_{ip}} 
        = \frac{1}{\tau}(P_{i,p} - 1)
\label{equ-gradient-sip}
\end{equation}

\begin{equation}
        g_{in}
        =\frac{\partial  L_{SupCon}^{i}}{\partial s_{\mathit{in}}} 
        = \frac{1}{\tau} P_{i,n}
\label{equ-gradient-sin}
\end{equation}

\begin{figure}[htbp]
  \centering
  \begin{minipage}[c]{0.55\textwidth}
    \centering
    \includegraphics[width=\linewidth]{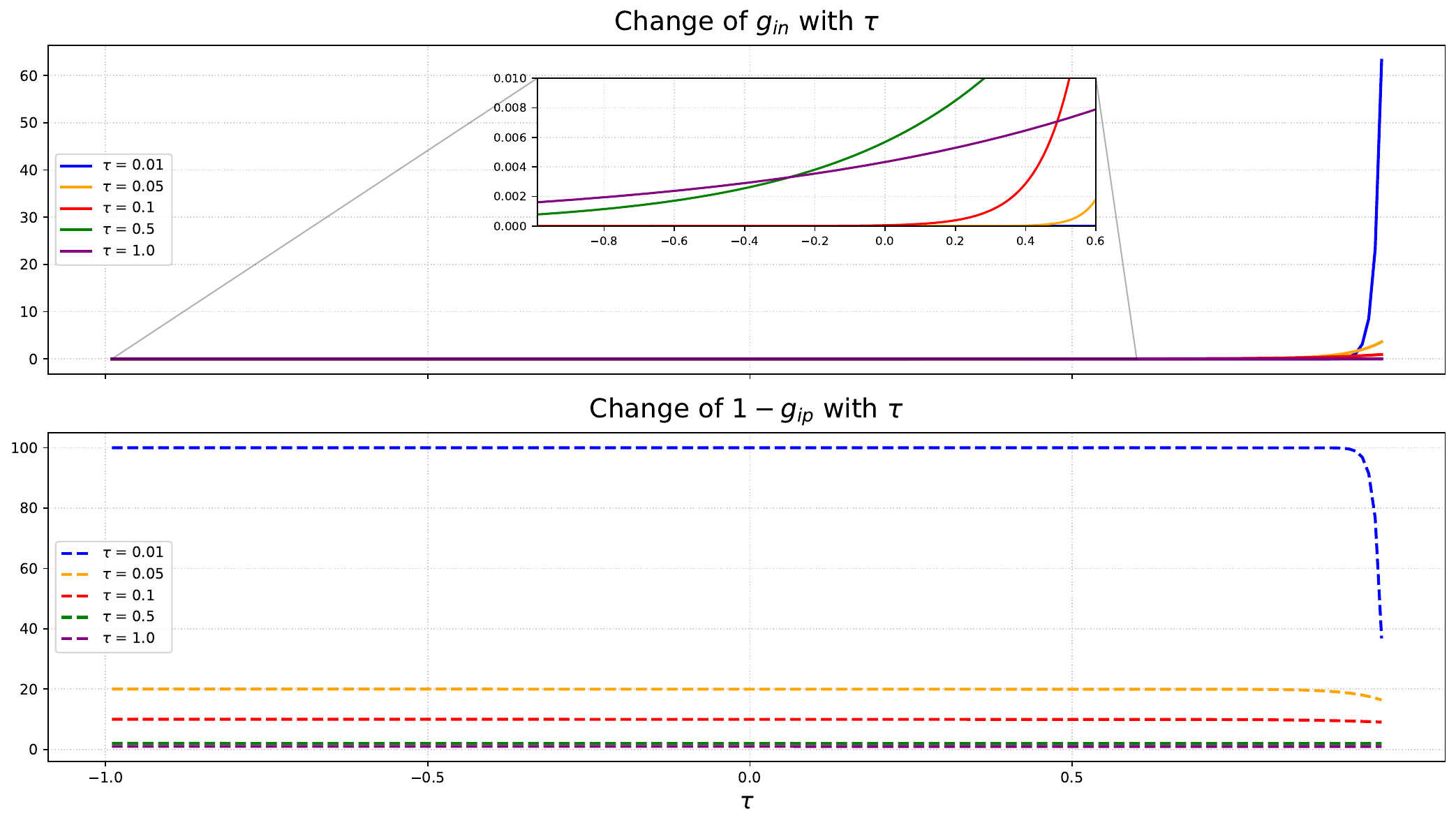}
    \vspace{1pt}
    \caption{Gram matrices of the class-wise representations on CIFAR10 and TinyImageNet datasets with the change of the temperature.}
    \label{fig-g-t}
  \end{minipage}
  \hspace{2pt}
  \begin{minipage}[c]{0.38\textwidth}
    \centering
    \includegraphics[width=\linewidth]{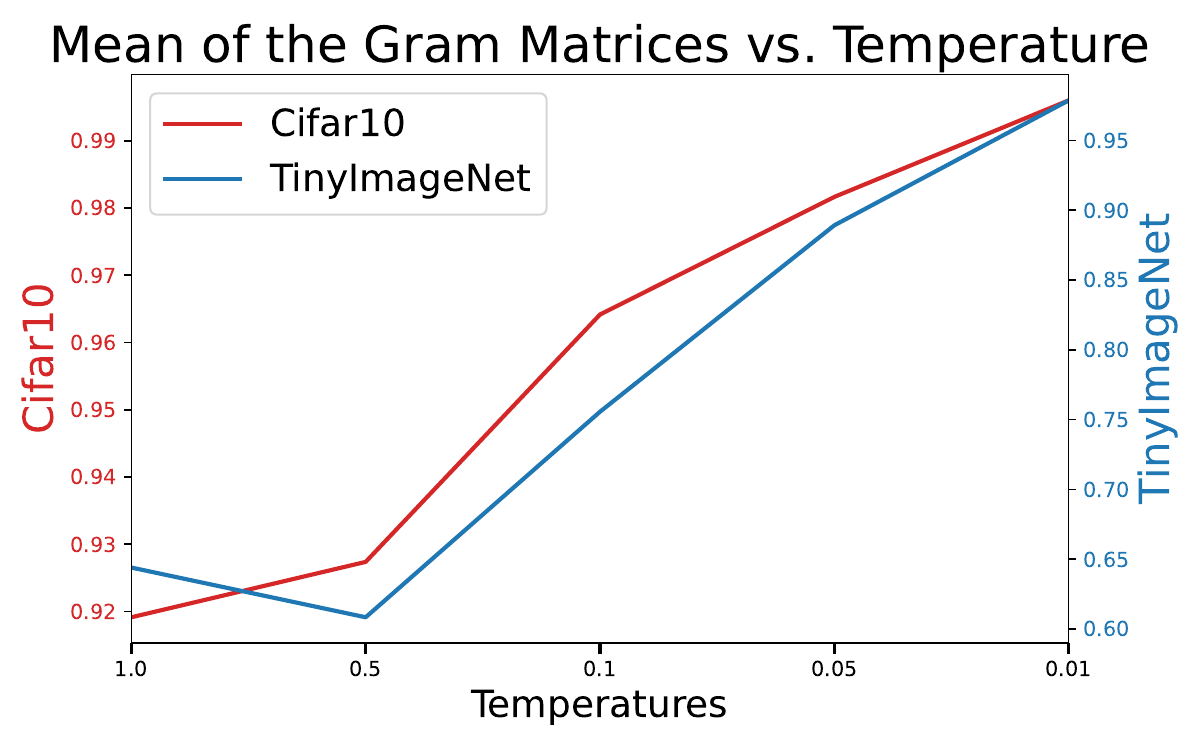}
    \vspace{1pt}
    \caption{Mean values of the Gram matrices of the representations of CIFAR10 and TinyImageNet models. The values increase with the decrease in temperature, indicating closer distances between the intra-class representations.}
    \label{fig-gram-value}
  \end{minipage}
\end{figure}

During training, Equations~\eqref{equ-gradient-sip} and~\eqref{equ-gradient-sin} show that the updates of $s_{ip}$ and $s_{in}$ are proportional to $|\frac{\partial  L_{SupCon}^{i}}{\partial s_{ip}}| = -g_{ip} = \frac{1}{\tau}(1 - P_{i,p})$ and $|\frac{\partial  L_{SupCon}^{i}}{\partial s_{\mathit{in}}}| = g_{in} =\frac{1}{\tau} P_{i,n}$ respectively, where $0 \leq P_{i,p}, P_{i,n} \leq 1$.
To examine the effect of the temperature parameter, we numerically compute and visualize $-g_{ip}$ and $g_{in}$ in Figure~\ref{fig-g-t}, with the similarity range $s_{ij}$ set to $[-1,1]$. The plots show that $\left|\frac{\partial L_{SupCon}^{i}}{\partial s_{ip}}\right|$ generally decreases as $\tau$ increases, indicating that smaller temperatures impose stronger attraction between positive pairs.
For negative pairs, consistent with the observations in self-supervised contrastive learning~\cite{wang2021understanding}, hard negatives with high similarity scores $s_{in}$ receive larger gradients under smaller temperatures, whereas larger temperatures assign relatively greater gradients to easy and moderately difficult negatives. Detailed mathematical analysis is provided in Appendix~\ref{app-p-t}.



Previous studies have shown that emphasizing hard negatives can improve representation quality in self-supervised contrastive learning~\cite{wang2021understanding,han2020self}. Based on the above analysis, this observation can be naturally extended to supervised contrastive learning.
For hard negatives, which typically share fewer readily distinguishable cues with the anchor samples, increasing their separation requires the model to capture more invariant and discriminative characteristics. This process encourages the learning of richer inter-class representations. Although the gradients associated with hard and easy positive pairs remain relatively similar for a fixed temperature $\tau$, smaller temperatures produce overall larger gradients for $s_{ip}$, leading to different optimization behaviors and feature learning dynamics.
These observations suggest that models trained with different temperatures may emphasize distinct aspects of the feature space.
Based on all the above, we hypothesize that \emph{using variant temperatures enables the learning of complementary representations}.

We compute class-wise Gram matrices using normalized representations. The off-diagonal entries measure pairwise cosine similarities among samples from the same class. As the temperature decreases, the mean similarity increases as shown in Figure~\ref{fig-gram-value}, indicating stronger positive-pair alignment and more compact intra-class representations. This observation is consistent with the gradient analysis, where smaller temperatures impose larger attraction gradients on positive pairs.
However, higher intra-class similarity alone does not necessarily imply richer feature diversity.
We then evaluate the prediction agreement between the models on novel data and prediction accuracy on in-set data after representation aggregation of variant models.
Since models trained with identical objectives and hyperparameters can still learn diverse representations~\cite{wang2018towards}, we take the models trained with the same temperature as a baseline for both evaluations.
We train SupCon models on the CIFAR-100 and TinyImageNet datasets using temperature values from ${1.0, 0.5, 0.1, 0.05, 0.01, 0.005}$ and then record their predictions on outlier samples and compute the agreement ratio between model pairs, defined as the proportion of samples receiving identical predictions. The results are summarized in Figure~\ref{fig-agreement}, while detailed numerical results are provided in Appendix~\ref{app-agreement}.
In Figure~\ref{fig-agreement}, the diagonal entries correspond to the agreement ratios between models trained with the same temperature. We observe that models trained with different temperatures generally exhibit lower agreement ratios than those trained with identical temperatures, indicating that varying $\tau$ leads to more diverse learned representations.
Besides, we evaluate the aggregated representations through linear probing on the in-distribution classes. Specifically, we concatenate representations extracted from models trained with identical and different temperatures. As shown in Figure~\ref{fig-pred}, representations aggregated from different-temperature models consistently achieve higher classification accuracy. The performance gain on CIFAR-10 is relatively small because the individual models already achieve very high accuracy ($97\%$ on average).
Overall, these results support our hypothesis that models trained with variant temperatures learn complementary representations.

\begin{figure}[htbp]
\centering 
\begin{minipage}[t]{0.48\textwidth} 
\centering
\includegraphics[width=\linewidth]{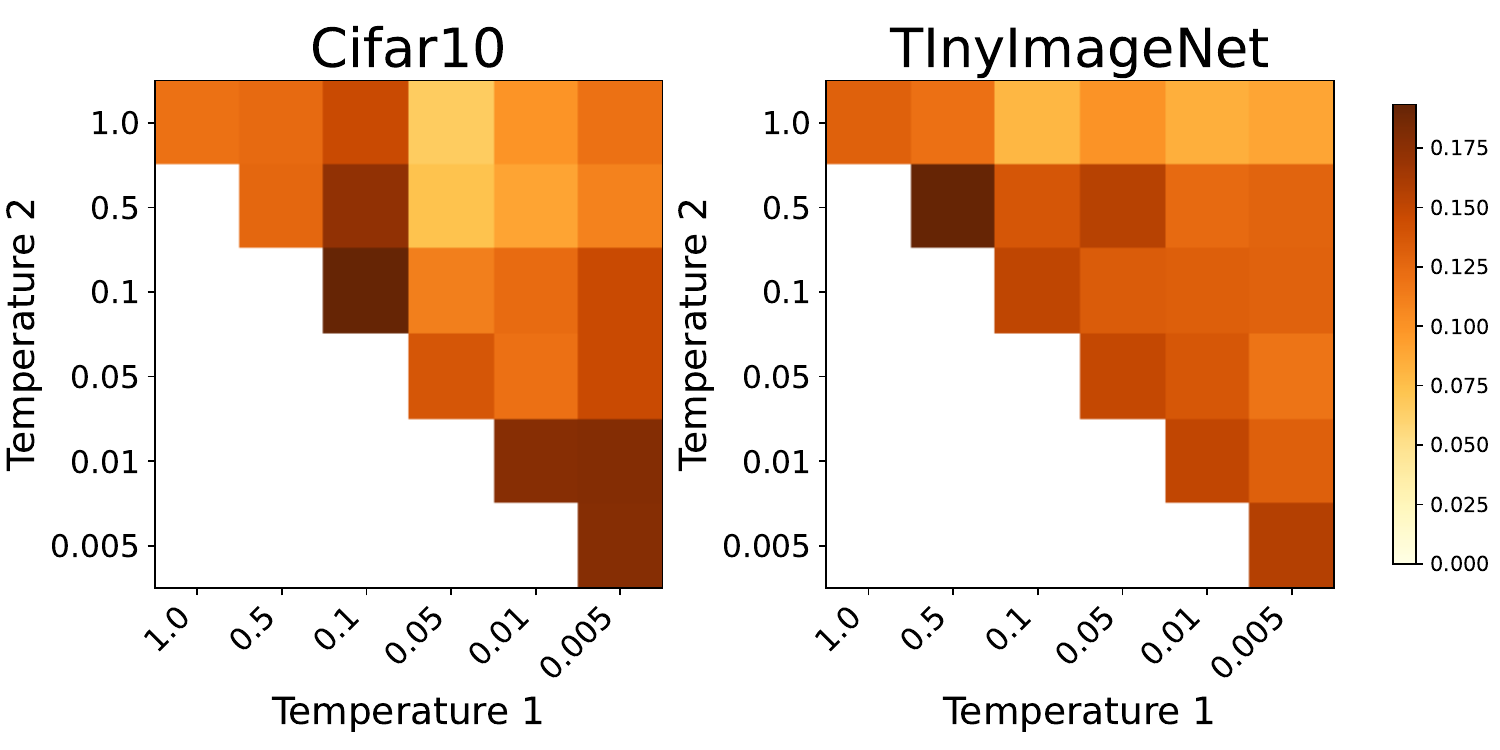}
\vspace{1pt}
\caption{Visualization of the prediction agreements between the models trained with the same and variant temperatures. Models with different temperatures demonstrate lower agreement. }
\label{fig-agreement}
\end{minipage}
\hfill
\begin{minipage}[t]{0.48\textwidth} 
\centering
\includegraphics[width=\linewidth]{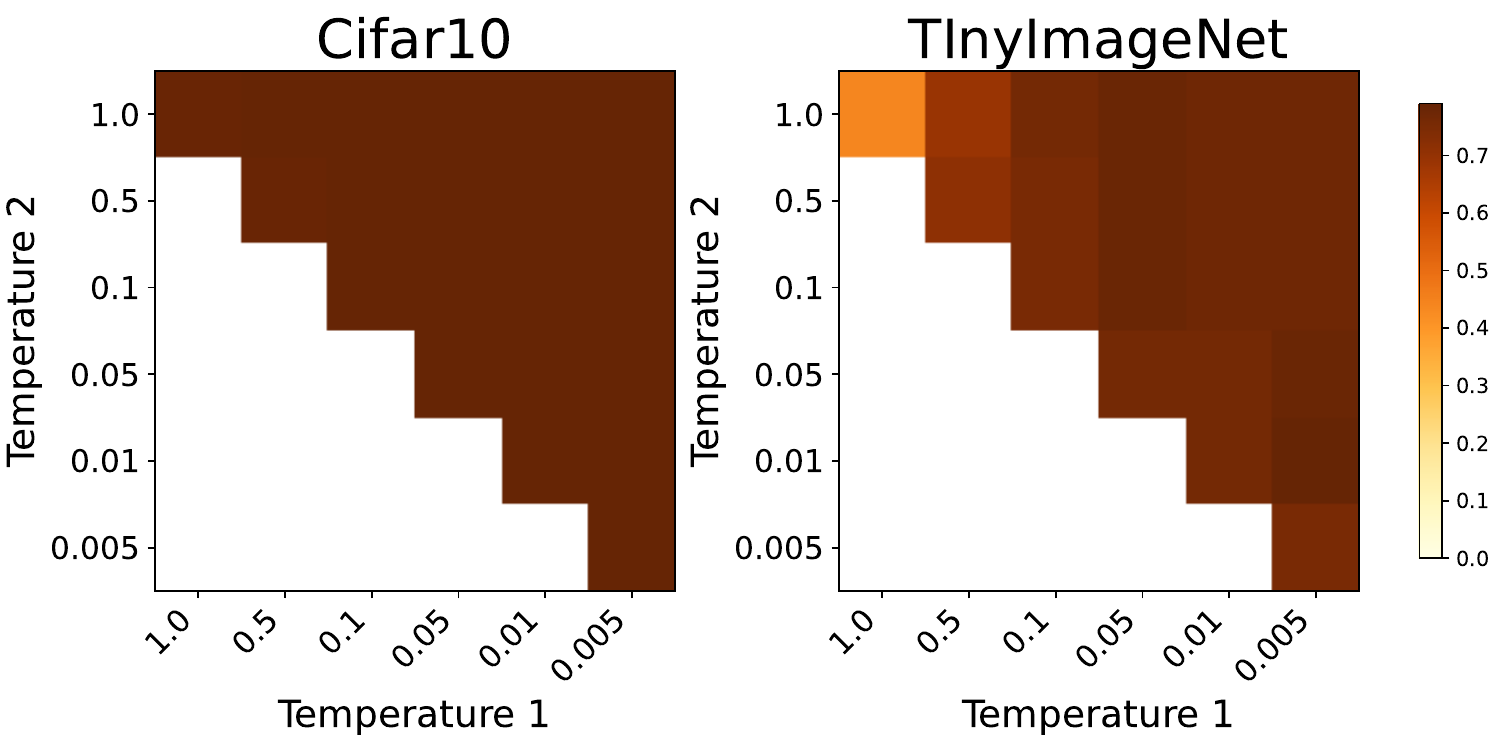}
\vspace{1pt}
\caption{Visualization of the prediction accuracy: the accuracy of the features with different temperatures is higher, indicating their richer semantic features.}
\label{fig-pred}
\end{minipage}
\end{figure}

\subsection{Feature Aggregation for OSR}
\begin{figure}
    \centering
    \includegraphics[width=0.8\linewidth]{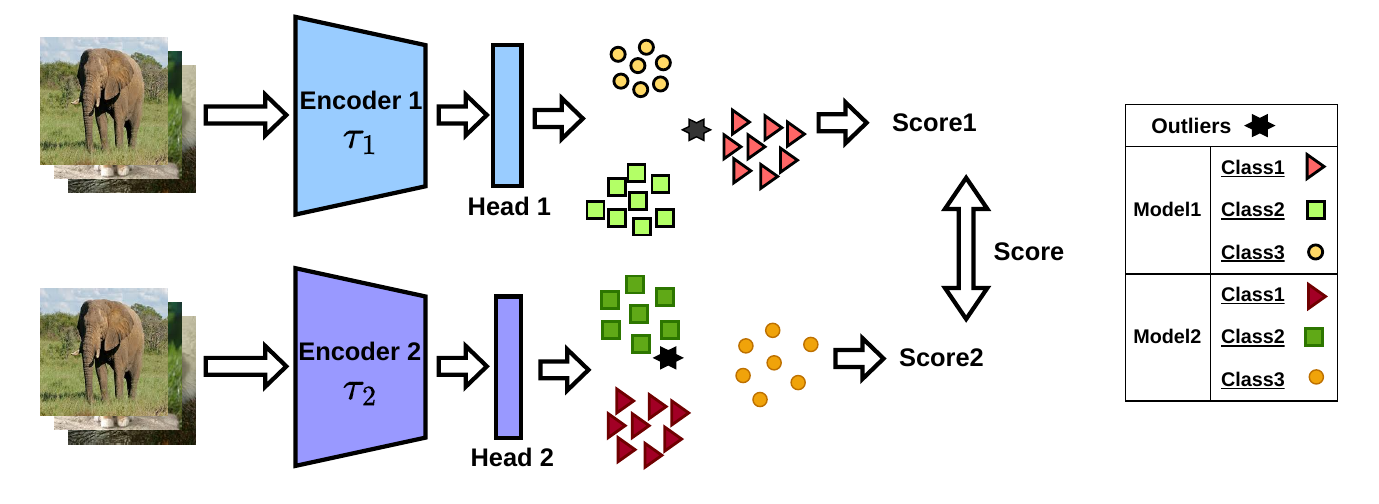}
    \vspace{3pt}
    \caption{Graphical illustration of our method (using two ensembling models as an example): SupCon models are trained with different temperatures, $\tau_1$ and $\tau_2$. The inlier representations demonstrate variant intra- and inter-class geometry.
    The OSR score of each ensembling representation is summed for aggregation.}
    \label{fig-method}
\end{figure}

Based on the above findings, we hypothesize that increasing the diversity of features encoded in the representations can improve open-set recognition performance. Our analysis further shows that supervised contrastive learning models trained with different temperature values tend to focus on different aspects of the data and consequently learn complementary representations.
Motivated by these observations, we propose an ensemble approach that leverages SupCon models trained with different temperatures, as illustrated in Figure~\ref{fig-method}. Specifically, we train multiple SupCon models independently using different temperature values and aggregate their OSR scores during inference. As will be shown in the following experiments, even this simple aggregation strategy consistently improves performance, supporting our hypothesis regarding the importance of feature diversity in OSR.

\section{Experiments} \label{sec-experiments}

\subsection{Settings} \label{subsec-exp-settings}
\noindent\textbf{Datasets} Following the OSR testbench in literature~\cite{vaze2022openset,arpl,neal2018open}, we evaluate our method on six split protocols, namely MNIST~\cite{deng2012mnist}, SVHN~\cite{goodfellow2013multi}, CIFAR10~\cite{cifar}, CIFAR+10~\cite{cifar}, CIFAR+50~\cite{cifar}, and TinyImagenet~\cite{deng2009imagenet}. The number of training classes (denoted using K) and the total number of testing classes (denoted using M), and their data sources of each protocol are in Appendix~\ref{app-data-protocol}. The complexity of each protocol is measured using \emph{openness}, $O=1 - \sqrt{K/M}$, which is the portion of outlier classes over the total number of classes. 

\noindent\textbf{Metrics} 
Since detecting open set samples requires setting the thresholds manually and a direct result comparison with different thresholds is not reasonable, a threshold-independent metric, the \emph{area under the receiver operating characteristic} (AUROC) curve~\cite{narkhede2018understanding} is one of the most commonly utilized metrics to evaluate the OSR methods.
In the AUROC curve, the true positive rate is plotted against the false positive rate by varying the threshold. The higher the AUROC is, the better the model can distinguish outliers from inliers. 

\noindent\textbf{Implementation details}
We use \emph{ResNet-18} \cite{he2016deep} as the backbone for all the models in this work without any pre-training or foundation models. The data augmentation strategies applied for contrastive learning are standard, i.e., random flip, color jitter, and gray scaling.
The hyperparameter settings are in Appendix \ref{app-hyperparamters}. Since we use representation norms as scores for OSR, which can not be applied for classification, we hence apply k-nearest-neighbor ($K=3$) with the concatenated representations for classifying inliers.
The influence of temperature on contrastive learning is highly nonlinear. Since $\tau$ appears in the denominator of the exponential term, changes at small temperatures have a much larger effect than equivalent absolute changes at large temperatures. Therefore, we sample temperatures approximately uniformly in logarithmic space.
The candidate temperatures are $\{1, 0.5, 0.1,  0.05, 0.01, 0.005\}$.
We take the aggregation of three representations for the final implementation. The final settings of the temperatures in the following experiments are in Table~\ref{tab-Temperatures} of the Appendix.

\subsection{Results} \label{sub-results}
The open set recognition (AUROC) results are in Table \ref{tab-auroc-results}. 
The baselines are the popular OSR methods, as well as the closest works to ours that introduced in Section~\ref{sec-background}.
We use GeoEnsemble and  Kodama-SupCon to denote the methods in~\cite{perera2021geometric} and~\cite{kodama2021open}. The rest of the abbreviations follow the original papers.
Our method can surpass most baselines and is comparable with SupCon-ST~\cite{xu2023contrastive}, which applies supervised contrastive learning as well.
Notably, our methods perform best on the most complex TinyImagenet dataset. We believe the reason lies in the complexity of the dataset, which provides a larger room for our method to exploit the variant aspects of the data. Figure~\ref{fig-cka} in Appendix~\ref{app-cka} demonstrates the CKA~\cite{kornblith2019similarity} values between the representations with variant temperatures, which infer similarities between the representations. Representations of the TinyImageNet dataset show smaller similarity with each other compared with CIFAR10, indicating richer features to learn that can support our point of view.

\begin{table*}[h]
\begin{center}
\small
\begin{tabular*}{0.99\textwidth}{c @{\extracolsep{\fill}} cccccccccc}
 \toprule
 &  \textbf{MNIST} & \textbf{SVHN} & \textbf{CIFAR10} & \textbf{CIFAR+10} & \textbf{CIFAR+50}  & \textbf{TinyImgNet} \\ 
 Openness & $22.54\%$ & $22.54\%$ & $22.54\%$ & $46.55\%$ & $72.78\%$ & $68.37\%$  \\ [0.5ex]  
 \hline
 Cross Entropy  &  $97.8 $  &  $88.6 $  &  $67.7 $  & $81.6 $  & $80.5 $   &  $ 57.7 $  \\
 Openmax  \cite{bendale2016towards} &  $98.1 $  &  $89.4$  &  $69.5 $  & $81.7 $  & $79.6$   &  $ 57.6 $  \\
 G-Openmax  \cite{GOpenMax_Ge17}       &  $98.4 $  &  $89.6 $  &  $67.5$  & $82.7$  & $81.9$   &  $ 58.0$  \\
 OSRCI  \cite{neal2018open}     &  $98.8$  &  $90.1$  &  $69.9$  & $83.8$  & $82.7$   &  $ 58.6 $  \\
 C2AE \cite{oza2019c2ae}        & $98.9 $  &  $92.2$  &  $89.5 $  & $95.5$  & $93.7$   &  $ 74.8$  \\
 APRL \cite{arpl}               & $99.6$  &  $96.3$  &  $90.1$  & $96.5$  & $94.3$   &  $ 76.2$  \\
 APRL-CS \cite{arpl}           & $\mathbf{99.7}$  &  $96.7$  &  $91.0 $  & $97.1$  & $95.1 $   &  $ 78.2$  \\
 OpenAUC \cite{wang2022openauc}  & $99.4$   & $95.0$  & $89.2$  & $95.2$ & $93.6$  &  $75.9$   \\ 
 GeoEnsemble \cite{perera2021geometric}  & $-$   & $95.8$  & $82.1$  & $93.7$ & $93.0$  &  $70.9$   \\
 Kodama-SupCon \cite{kodama2021open}  & $-$   & $95.5$  & $84.2$   & $95$  & $94.6$&  $77$   \\ [0.5ex] 
SupCon-ST \cite{xu2023contrastive}  & $99.7$   & $\mathbf{99.1}$  & $\mathbf{94.2}$  & $\mathbf{98.1}$ & $\mathbf{97.3}$  &  $80.9$ \\ 
MEDAF\cite{wang2024exploring} & $-$   & $95.7$  & $86$  & $96$ & $95.5$  &  $80$   \\ [0.5ex] 
\hline
Ours  & $\mathbf{99.7}$   & $97.71$   & $93.12$  & $\mathbf{98.1} $ & $96.11$ & $\mathbf{81}$   \\
 \bottomrule
\end{tabular*}
\vspace{5pt}
\caption{The area under the ROC curve (AUROC) (in $\%$) for detecting known and unknown samples (Results of the baseline methods are from the original papers). 
Bold numbers indicate the best results. Our method can achieve the best or comparable results in most testing protocols, especially on TinyImagenet.}
\label{tab-auroc-results}
\end{center}
\end{table*}

\subsection{Ablation Study} \label{abl-aggre}

\subsubsection{Single versus Aggregation}
To illustrate the effectiveness of the representation aggregation, we compare the OSR performance when using single, double, and triple representations with different temperatures. When comparing with the results in Table~\ref{tab-temperatures}, Table~\ref{tab-temperatures-single}, and Table~\ref{tab-temperatures-two}. The OSR performance with aggregation of three representations demonstrates the best (with the average AUROC change from $89.25\%$, $91.02\%$ to $92.43\%$ for CIFAR10, and from  $74.36\%$, $78.75\%$ to $79.57\%$ for TinyImagenet). The representation aggregation is more effective than a single representation.

\begin{table}[h]
\centering
\begin{minipage}{0.3\linewidth}
\centering
\scriptsize
\begin{tabular}{c|cc}
    \toprule
         &	CIFAR10	& TinyImgnet	\\
         \hline
     1.0   & $80.17$ &$58.57$	 \\
     0.5    & $88.96$ &$78.52$	 \\
     0.1 & $90.69$  &$75.06$	 \\
     0.05 & $91.2$ & $78.46$	 \\
     0.01 & $91.02$ &$76.54$	 \\
     0.005 & $91.75$ &$77.04$	 \\
     Avg & $89.25$ &$74.36$  \\
     \bottomrule
    \end{tabular}
\vspace{5pt}
\captionof{table}{The AUROC (in $\%$) results when using single representation with variant temperatures.}
\label{tab-temperatures-single}
\end{minipage}
\hfill
\begin{minipage}{0.3\linewidth}
\centering
\scriptsize
\begin{tabular}{c|cc}
    \toprule
         &	CIFAR10	& TinyImgnet	\\
         \hline
      0.5, 0.1     & $91.84$ &$78.99$	 \\
     0.5, 0.05    & $92.82$ &$81.42$	 \\
     0.5, 0.01    & $92.75$ &$79.36$	 \\
     0.1, 0.05    & $92.79$ &$79.33$	 \\
     0.1, 0.01    & $92.83$ &$78.66$	 \\
     0.1, 0.005   & $92.69$ &$77.86$	 \\
     Avg      & $91.02$ &$78.75$  \\
     \bottomrule
    \end{tabular}
\vspace{5pt}
\captionof{table}{The AUROC (in $\%$)results when aggregating two representation with variant temperatures. Whole results are Table~\ref{tab-2} in Appendix~\ref{app-ab}.}
\label{tab-temperatures-two}
\end{minipage}
\hfill
\begin{minipage}{0.32\linewidth}
\centering
\scriptsize
\begin{tabular}{c|cc}
    \toprule
         &	CIFAR10	& TinyImgnet	\\
         \hline
     RepCat    & $93.08$ &$79.62$	 \\
     RepSum     & $93.1$ &$79.43$	 \\
     SocSum    & $\mathbf{93.23}$ &$\mathbf{79.97}$	 \\
     \bottomrule
    \end{tabular}
\vspace{5pt}
\caption{The AUROC (in $\%$) results when using different aggregation strategies. All results are the average over all different temperature combinations in Table~\ref{tab-temperatures}. \emph{SocSum} can show the best performance.}
    \label{tab-aggregation}
\end{minipage}
\end{table}

\begin{table}[h]
    \centering
    \scriptsize
    \begin{tabular}{c|c|c|c|c|c|c|c}
    \toprule
         & 0.5,0.05,0.005 &	0.5,0.1,0.05	& 0.5,0.1,0.01 & 0.5,0.1,0.005	& 0.1,0.05,0.01 & 0.1,0.05,0.005 & Avg  \\
         \hline
     CIFAR10    & $93.04$ &$92.96$	& $93.06$ & $92.9$  &	$\mathbf{93.12}$ &  $93$ & $92.43$ \\
     \hline
    TinyImgnet & $80.3$ & $\mathbf{81}$	& $78.8$ & $79.48$ & $79.14$	&  $79.68$  &  $79.57$ \\
     \bottomrule
    \end{tabular}
    \vspace{5pt}
    \caption{Comparison of the AUROC (in $\%$) results when aggregating three representations with variant temperatures. Whole results are in Table~\ref{tab-3} in Appendix~\ref{app-ab}.}
    \label{tab-temperatures}
\end{table}

\subsubsection{The Choice of the Temperatures}
We compare the OSR performances when aggregating the representations with variant combinations of the temperatures.
The results in Table~\ref{tab-temperatures} show that a different combination of temperatures can output different OSR performances. It aligns with our analysis in~\ref{subsubsec-temp-feature} that the attentions on the features of SupCon vary with the temperatures and the similarities between the representations, which differ between datasets.

\subsubsection{Aggregation Strategies} \label{abl-aggre}
We compare the OSR performances using different aggregating strategies. Assume that the representations to aggregate are $\mathbf{Z}=\{\mathbf{z}_1$, $\mathbf{z}_2$,..., $\mathbf{z}_i\}$, and we consider three aggregation strategies, which are representation concatenating (\emph{RepCat}), representation summation (\emph{RepSum}), and score summation (\emph{SocSum}). 
In \emph{RepCat}, a super representation $\mathbf{z}_{s}$ is the concatenation of $\mathbf{z}_i$s, i.e., $\mathbf{z}_s=\mathbf{z}_1 \oplus \mathbf{z}_2... \oplus \mathbf{z}_i$, and then the OSR scores are computed on $\mathbf{z}_{s}$. 
In \emph{RepSum}, a dimension-wise mean representation $\mathbf{z}_{m}$ is first computed over all $\mathbf{z}_i$s, i.e., $\mathbf{z}_m = \frac{\sum\limits_{i}\mathbf{z}_i}{|\mathbf{Z}|}$, and then the score is calculated with $\mathbf{z}_{m}$.
In \emph{SocSum}, the OSR scores are firstly computed with each $\mathbf{z}_i$, and then summed together. We compare their OSR performances on CIFAR10 and TinyImagenet protocols in Table~\ref{tab-aggregation}, which are the average over all different temperature configurations in Table~\ref{tab-temperatures}. \emph{SocSum} can demonstrate the highest performance, although the differences are not significant. We think the reasons lie in the fact that SocSum can be more robust to the noises and redundancies in the representations.

\section{Conclusion \& Future Work}

In this work, we investigate the problem of open set recognition. Through a series of controlled experiments, we demonstrate a positive correlation between OSR performance and the diversity of features encoded in learned representations. We further analyze supervised contrastive learning from the perspective of the temperature parameter, showing that different temperatures alter the model's emphasis on positive and negative sample pairs, thereby leading to representations that capture complementary features.
Motivated by this observation, we propose a simple yet effective approach that aggregates representations learned by supervised contrastive models trained with different temperatures for OSR. Extensive experiments on standard OSR benchmarks demonstrate that the proposed method achieves competitive or state-of-the-art performance while requiring only minimal modifications to existing supervised contrastive learning frameworks.

Despite the promising results, several open questions remain. First, our controlled experiments indicate that shape-related features are not entirely suppressed even when they are not strictly required for the classification task. This observation suggests that developing methods to encourage the encoding of a broader range of auxiliary features is a worthwhile direction. For example, prior work has explored incorporating self-supervised objectives alongside supervised losses for this goal~\cite{lee2025theoretical}.
Second, studies have reported that representations from earlier or intermediate layers can exhibit richer features~\cite{marczak2024revisiting}, which is consistent with our experimental findings. This motivates further investigation into how representations from multiple layers can be leveraged for more effective outlier detection.
Finally, we observe substantial performance variations across different outlier detection methods, which we believe are strongly influenced by the characteristics of the underlying deep representations. This highlights the need for a systematic evaluation of the detection techniques across diverse representation learning paradigms.

An interesting direction for future research is to develop methods that enable a single model to learn equally diverse or even more expressive representations, thereby retaining the benefits of representation diversity while reducing the computational overhead associated with model ensembles.
Furthermore, although this work focuses on open set recognition, the insights obtained from our analysis are not limited to this setting. Since the proposed framework is fundamentally based on representation diversity and complementary feature learning, we believe that these principles may also benefit other tasks, including transfer learning, model robustness, and domain adaptation. Exploring these directions constitutes promising future work.
%
%
%
\bibliographystyle{splncs04}
\bibliography{mybibliography}

\clearpage
\setcounter{page}{1}
\appendix

\section{Details for \ref{sec-toy-example}} \label{app-toy-example}

\subsection{Network Architecture}  \label{app-toy-archicture}

We trained multi-layer perceptrons for E1 and E2 in Section \ref{sec-toy-example}. The network architecture is in Table \ref{tab-toy-architecture} and the code is online available\footnote{https://github.com/gawainxu/Data\_Bias\_Experiments}, The three-dimensional input/output sizes follow the format of height$\times$width$\times$channels.

\begin{table}[h]
\centering
\begin{tabular}{ c|cccc } 
 \toprule
Name & Type & Input & Output & Kernel Size \\
 \midrule
conv1 & Conv2D & 64$\times$64$\times$3 & 64$\times$64$\times$10 & 5$\times$5$\times$10 \\ 
avgpool & AvgPooling2D & 64$\times$64$\times$10  &  32$\times$32$\times$10 &  -   \\
flatten & Flatten   &  32$\times$32$\times$10 &    10240  &  - \\
linear1 & Linear  &   10240  & 1000   & - \\
linear2 & Linear  &   1000  & 20   & - \\
linear3 & Linear  &   20  & num-classes & -   \\
 \bottomrule
\end{tabular}
\vspace{6pt}
\caption{Network Architectures for E1 and E2 in Section \ref{sec-toy-example}.}
\label{tab-toy-architecture}
\end{table}

\subsection{Setting for the Cifar100 Dataset} \label{app-cifar100}

The splitting of the groups and the corresponding class names are in Table~\ref{tab-cifar100-group}, which follows the introduction in~\cite{cifar}.

\begin{table}[h]
\centering
\begin{tabular}{ c|ccccc} 
 \toprule
Groups & 1 & 2 & 3 & 4 & 5 \\
 \midrule
& \makecell{beaver, \\ aquarium fish, \\ orchids, \\ bottles, \\ apples, \\ clock, \\ bed, \\ bee, \\ bear, \\ bridge, \\ cloud, \\ camel, \\ crab, \\ baby, \\ crocodile, \\ hamster, \\ maple, \\ bicycle, \\ lawn-mower}  & \makecell{dolphin, \\ flatfish, \\ poppies, \\ bowls, \\ mushrooms, \\ keyboard, \\ chair, \\ beetle, \\ leopard, \\ castle, \\ forest, \\ cattle, \\ lobster, \\ boy, \\ dinosaur, \\ mouse, \\ oak, \\ bus, \\ rocket} & \makecell{otter, \\ ray, \\ roses, \\ cans, \\ oranges, \\ lamp, \\ couch, \\ butterfly, \\ lion, \\ house, \\ mountain,\\  chimpanzee \\ snail, \\ girl, \\ lizard, \\ rabbit, \\ palm, \\ motorcycle, \\ street car} & \makecell{seal, \\ shark, \\ sunflowers, \\ cups, \\ pears, \\ telephone, \\ table, \\ caterpillar, \\ tiger, \\ road, \\ elephant, \\ raccoon, \\spider, \\ man, \\ snake, shrew, \\ pine, \\ truck, \\ tank} & \makecell{whale, \\ trout, \\ tulips, \\ plates, \\ peppers, \\ television, \\ wardrobe, \\ cockroach, \\ wolf, \\ skyscraper, \\ kangaroo, \\ skunk, \\ worm, \\ woman, \\ turtle, \\ squirrel, \\ willow, \\ train, \\ tractor} \\ 
 \bottomrule
\end{tabular}
\vspace{6pt}
\caption{Class names in each group of the CIFAR100 dataset}
\label{tab-cifar100-group}
\end{table}

\section{Derivation of $\frac{\partial L_{SupCon}^{i}}{\partial s_{in}}$ and $\frac{\partial L_{SupCon}^{i}}{\partial s_{ip}}$} \label{app-gradients}

Following Section \ref{sec-method-pre}:

\begin{align*}
    L_{SupCon}^{i} &= -\sum\limits_{p \in P(i)}\log \frac{\exp (\frac{s_{ip}}{\tau})}{\sum\limits_{\substack{j \neq p,n \\ j \in A(i)}} \exp(\frac{s_{ij}}{\tau}) + \exp (\frac{s_{ip}}{\tau}) + \exp (\frac{s_{in}}{\tau})} \\
    &= -\sum\limits_{p \in P(i)} [\frac{s_{ip}}{\tau} - \log (\sum\limits_{\substack{j \neq p,n \\ j \in A(i)}} \exp(\frac{s_{ij}}{\tau}) + \exp (\frac{s_{ip}}{\tau}) + \exp (\frac{s_{in}}{\tau}))]
\end{align*}

For one given negative sample $n$ corresponding to anchor sample $i$,

\begin{align*}
    \frac{\partial  L_{SupCon}^{i}}{\partial s_{in}} &= \frac{\partial \log (\sum\limits_{\substack{j \neq p,n \\ j \in A(i)}} \exp(\frac{s_{ij}}{\tau}) + \exp (\frac{s_{ip}}{\tau}) + \exp (\frac{s_{in}}{\tau}))}{\partial s_{in}} \\
    &= \frac{\frac{1}{\tau}\exp (\frac{s_{in}}{\tau})}{\sum\limits_{\substack{j \neq p,n \\ j \in A(i)}} \exp(\frac{s_{ij}}{\tau}) + \exp (\frac{s_{ip}}{\tau}) + \exp (\frac{s_{in}}{\tau})}  \\
    &= \frac{1}{\tau} P_{i,n}
\end{align*}

As defined in~\ref{sec-method-pre}, $P_{i,n} = \frac{\exp (\frac{s_{in}}{\tau})}{\sum\limits_{\substack{j \neq p,n \\ j \in A(i)}} \exp(\frac{s_{ij}}{\tau}) + \exp (\frac{s_{ip}}{\tau}) + \exp (\frac{s_{in}}{\tau})}$, which is the softmax value of $\frac{s_{in}}{\tau}$.
Similarly, for one given positive sample $p$,

\begin{align*}
    \frac{\partial  L_{SupCon}^{i}}{\partial s_{ip}} &= - \frac{\partial [\frac{s_{ip}}{\tau} - \log (\sum\limits_{\substack{j \neq p,n \\ j \in A(i)}} \exp(\frac{s_{ij}}{\tau}) + \exp (\frac{s_{ip}}{\tau}) + \exp (\frac{s_{in}}{\tau}))]}{\partial s_{ip}} \\
    &= - [\frac{1}{\tau} - \frac{\frac{1}{\tau}\exp (\frac{s_{ip}}{\tau})}{\sum\limits_{\substack{j \neq p,n \\ j \in A(i)}} \exp(\frac{s_{ij}}{\tau}) + \exp (\frac{s_{ip}}{\tau}) + \exp (\frac{s_{in}}{\tau})}] \\
    &= \frac{1}{\tau}(P_{i,p} - 1)  
\end{align*}

\section{Relations between $|\frac{\partial  L_{SupCon}^{i}}{\partial s_{ip}}|$ and $|\frac{\partial  L_{SupCon}^{i}}{\partial s_{\mathit{in}}}|$ and $\tau$} \label{app-p-t}

In order to look into the relations between $|\frac{\partial  L_{SupCon}^{i}}{\partial s_{ip}}|$ and $|\frac{\partial  L_{SupCon}^{i}}{\partial s_{\mathit{in}}}|$ and $\tau$, we compute their first derivation. With 

\begin{equation*}
        |\frac{\partial  L_{SupCon}^{i}}{\partial s_{ip}}|
        = \frac{1}{\tau}(1 - P_{i,p}), 
        |\frac{\partial  L_{SupCon}^{i}}{\partial s_{\mathit{in}}}|
        = \frac{1}{\tau} P_{i,n}
\end{equation*}

and we set

\begin{align*}
    \frac{\partial P_{i,j}}{\partial \tau} &= \frac
    {\partial \frac{\exp(\frac{s_{ij}}{\tau})} {\sum\limits_{a\in A}\frac{\exp(s_{ia})}{\tau}}}
    {\partial\tau}
    = \frac{-\frac{s_{ij}}{\tau^2}\exp(\frac{s_ij}{\tau})\sum\limits_{a\in A}\exp(\frac{s_{ia}}{\tau})+\exp(\frac{s_{ij}}{\tau})\sum\limits_{a\in A}\frac{s_{ia}}{\tau^2}\exp(\frac{s_{ia}}{\tau})} 
    {({\sum\limits_{a\in A}\exp(\frac{s_{ia}}{\tau})})^{2}} \\
    &= \frac{1}{\tau^2}(-s_{ij}P_{i,j} + \frac{\sum\limits_{a\in A}s_{ia}\exp(\frac{s_{ia}}{\tau})}{\sum\limits_{a\in A}\exp(\frac{s_{ia}}{\tau})}P_{i,j})\\
    & = \frac{P_{i,j}}{\tau^2}(-s_{ij} + \bar{s_{ia}})
\end{align*}

with $\bar{s_{ia}} = \frac{\sum\limits_{a\in A}s_{ia}\exp(\frac{s_{ia}}{\tau})}{\sum\limits_{a\in A}\exp(\frac{s_{ia}}{\tau})}$, which is the \emph{Softmax} weighted average of $P_{i,a}$,
then

\begin{align*}
    \frac{\partial|\frac{\partial  L_{SupCon}^{i}}{\partial s_{ip}}|}{\partial \tau} &= 
    \frac{\partial \frac{1}{\tau} (1-P_{i,p})}{\partial \tau} \\
    &= -\frac{1}{\tau^2}(1-P_{i,p}) - \frac{1}{\tau} \frac{\partial P_{i,p}}{\partial \tau} \\
    &=  -\frac{1}{\tau^2}(1-P_{i,p}) + \frac{P_{i,p}}{\tau^3}(s_{ip} - \bar{s_{ia}})
\end{align*}

and

\begin{align*}
    \frac{\partial|\frac{\partial  L_{SupCon}^{i}}{\partial s_{in}}|}{\partial \tau} &= 
    \frac{\partial \frac{1}{\tau}  P_{i,n}}{\partial \tau} \\
    &= -\frac{1}{\tau^2}P_{i,n} +\frac{P_{i,n}}{\tau^3}(-s_{in} + \bar{s_{ia}}) \\
    &= \frac{P_{i,n}}{\tau^3}(-s_{in} + \bar{s_{ia}}-\tau)
\end{align*}

So far, to deduce the monotonicity of  $|\frac{\partial  L_{SupCon}^{i}}{\partial s_{ip}}|$ and $|\frac{\partial  L_{SupCon}^{i}}{\partial s_{\mathit{in}}}|$ with $\tau$, we first compare $s_{ip}$, $s_{in}$, and $\bar{s_{ia}}$. As defined, $\bar{s_{ia}} = \frac{\sum\limits_{a\in A}s_{ia}\exp(\frac{s_{ia}}{\tau})}{\sum\limits_{a\in A}\exp(\frac{s_{ia}}{\tau})}=\sum\limits_{a\in A}\text{Softmax}(s_{ia})s_{ia}$, then ${\min\limits_a\{s_{ia}\}} \leq \bar{s_{ia}} \leq \max\limits_a\{s_{ia}\}$. 
And $\text{Softmax}(s_{ia})$ increases with $s_{ia}$ and is dominated by positives and hard negatives. 
For easy negatives, the smaller $s_{in}$ is, the more possible that $s_{in} \ll \bar{s_{ia}}$, $-s_{in} + \bar{s_{ia}}-\tau > 0$, which is identical with the observation from Figure~\ref{fig-g-t} that $|\frac{\partial  L_{SupCon}^{i}}{\partial s_{\mathit{in}}}|$ increases with $\tau$.

It is infeasible to analysis for $s_{ip}$ from $\frac{\partial|\frac{\partial  L_{SupCon}^{i}}{\partial s_{ip}}|}{\partial \tau}$, we then divide the numerator and denominator of $P_{i,j}$ by $\exp(\frac{s_{max}}{\tau})$, where $s_{max}$ is the maximal value that $s_{ij}$ can be, i.e.,

\begin{equation*}
    P_{i,j} = \frac{\exp(\frac{s_{ij}}{\tau})/\exp(\frac{s_{max}}{\tau})}{(\sum\limits_{a\in A(i)}\exp(\frac{s_{ia}}{\tau}))/\exp(\frac{s_{max}}{\tau})} = \frac{\exp(\frac{s_{ij}-s_{max}}{\tau})}{\sum\limits_{a\in A(i)}\exp(\frac{s_{ia}-s_{max}}{\tau})}
\end{equation*}

When $s_{ij}$ is close to $s_{max}$, which is the case for most positives, then $s_{ij}-s_{max} \rightarrow 0$, $\exp(\frac{s_{ij}-s_{max}}{\tau}) \approx \exp(0)=1$. And $\sum\limits_{a\in A(i)}\exp(\frac{s_{ia}-s_{max}}{\tau}) \approx M$, $M$ is the number of $s_{ia}$ whose value is close to $s_{max}$. $1-P_{ip}$ is close to the constant of $1-\frac{1}{M}$. Hence, $\frac{\partial|\frac{\partial  L_{SupCon}^{i}}{\partial s_{ip}}|}{\partial \tau} \approx \frac{-1}{\tau^2}(1-\frac{1}{M}) < 0$, the changing rate of $|\frac{\partial  L_{SupCon}^{i}}{\partial s_{ip}}|$ decreases with $\tau$. The same analysis applies to hard negatives. Since $P_{in}$ approaches to constants, then $\frac{\partial|\frac{\partial  L_{SupCon}^{i}}{\partial s_{in}}|}{\partial \tau} \approx \frac{-1}{M\tau^2} < 0$, which means the changing rate of $|\frac{\partial  L_{SupCon}^{i}}{\partial s_{in}}|$ decreases with $\tau$ for hard negatives.



Therefore, for positives, $|\frac{\partial  L_{SupCon}^{i}}{\partial s_{ip}}|$ decreases with $\tau$. 
For easy negatives, when $\tau$ is small, $|\frac{\partial  L_{SupCon}^{i}}{\partial s_{in}}|$ increases with $\tau$.
The above analysis is consistent with the numerical results in~\ref{subsubsec-temp-feature}.

\section{Details for Prediction Agreement in Section~\ref{subsubsec-temp-feature}} \label{app-agreement}

As introduced in Section~\ref{subsubsec-temp-feature}, we train models on CIFAR10 and TinyImageNet datasets with variant temperatures, $[1.0, 0.5, 0.1, 0.05, 0.01, 0.005]$. The models are then evaluated on novel data. The dataset settings are in Table~\ref{tab-app-agreement}. All models are of the ResNet18 architecture. 

\begin{table}[h]
    \centering
    \begin{tabular}{c|c}
    \toprule
      Training Data   & Testing Data \\
    \midrule
    CIFAR10           & CIFAR100     \\
    20 classes in TinyImageNet   &   The rest classes in TinyImageNet  \\
    \makecell{[108, 147, 17, 58, 193, 123, 72, 144, \\ 75, 167, 134, 14, 81, 171, 44, \\ 197, 152, 66, 1, 133]}   & \makecell{[ 0,   1,   2,   3,   4,   5,   6,   7,   8,  \\
                           9,  10,  11,  12, 13,  14,  15,  16,  17, \\ 20,  
                           21,  22,  24,  25,  26,  27,  28, 30, \\  31,  32,  
                           33,  34,  36,  37,  39,  40,  41, \\ 42,  43,  44, 
                           45,  46,  47,  48,  49,  50, \\ 51,  52,  53,  54, 
                           55,  56,  58, 60,  62, \\ 63,  64,  65,  66,  68,  
                           69,  70,  71,  72, \\ 73,  74, 75,  76,  77,  78, 
                           79,  80,  81,\\  82,  83,  84,  86,  87,  88, 89, 
                           92,  93, \\ 94,  95,  96,  97,  98,  99, 100, 101, 
                           102,\\ 103, 104, 105, 106, 107, 108, 109, \\ 110, 111,
                           112, 113, 114, 115, 116, 117,\\ 118, 119, 120, 121, 
                           122, 123, 124, 125, 126, \\ 127, 128, 129, 130, 131, 
                           132, 135, 136, 137, \\ 138, 139, 140, 141, 142, 143, 
                           144, 145, 146, \\ 147, 148, 149, 150, 151, 152, 153, 
                           154, 155, \\ 156, 157, 158, 159, 160, 162, 163, 164, 
                           165, \\ 166, 167, 168, 169, 170, 171, 172, 173, 174, 
                           \\ 175, 176, 177, 178, 179, 180, 181, 182, \\ 183, 185, 
                           187, 188, 189, 190, 191, 192, \\ 193, 194, 195, 196, 197]}    \\
    \bottomrule
    \end{tabular}
    \vspace{6pt}
    \caption{Dataset settings for the prediction agreement evaluations in Section~\ref{subsubsec-temp-feature}.}
    \label{tab-app-agreement}
\end{table}

\section{Experimental Details for Section~\ref{sec-experiments}} \label{app-exp}

\subsection{Testing Protocol} \label{app-data-protocol}

The settings for the number of inlier and outlier classes and their source datasets are listed in Figure~\ref{tab-datasets-split}.

\begin{table}[h]
    \centering
    \begin{tabular}{ccc}
    \toprule[1.5pt]
        Protocols & \makecell{Known \\ ($\sharp$ Classes / Source)} & \makecell{Unknown \\ ($\sharp$  Classes / Source)}   \\
        \hline
        MNIST         &   6 / MNIST    &   4 / MNIST                \\
        SVHN          &   6 / SVHN     &   4 / SVHN             \\
        CIFAR10       &   6 / CIFAR10   &  4 / CIFAR10             \\
        CIFAR+10      &   4 / CIFAR10    &  10 / CIFAR100              \\
        CIFAR+50      &   4 / CIFAR10   &     50 / CIFAR100           \\
        TinyImagenet   & 20 / TinyImagenet     &   180 / TinyImagenet         \\
    \bottomrule[1.5pt]
    \end{tabular}
    \vspace{5pt}
    \caption{Datasets splitting protocols for known and unknown classes}
    \label{tab-datasets-split}
\end{table}

\subsection{Hyper-Parameters} \label{app-hyperparamters}
We list the hyperparameters used for training in Section \ref{sec-experiments} in Table \ref{tab-hyperparameter}. lr, epochs, and bz stand for learning rate, number of epochs, and batch size, respectively. The settings are for all models trained on each dataset.
We use \emph{Adam} optimizer \cite{kingma2014adam} with momentum to train the models, which is also standard in many contrastive learning works. 

\begin{table}[h]
    \centering
    \begin{tabular}{cccc}
    \toprule[1.5pt]
        Protocols & lr & epochs & bz  \\
        \hline
        MNIST         &   0.01   & 100  &  256 \\
        SVHN          &   0.01   & 300  &  256  \\
        CIFAR10       &   0.01   & 400  &  256 \\
        CIFAR+10      &   0.01   & 400  &  256 \\
        CIFAR+50      &   0.01   & 400  &  256 \\
        TinyImagenet   &  0.01   & 400  &  256 \\
    \bottomrule[1.5pt]
    \end{tabular}
    \vspace{5pt}
    \caption{Hyperparameters used for training in Section \ref{sec-experiments}.}
    \label{tab-hyperparameter}
\end{table}

\begin{table}[h]
    \centering
    \begin{tabular}{cc}
    \toprule[1.5pt]
        Protocols & Temperatures \\
        \hline
        MNIST         &  0.5, 0.1  \\
        SVHN          &  0.5, 0.1  \\
        CIFAR10       &  0.1, 0.05, 0.01  \\
        CIFAR+10      & 0.1, 0.05, 0.01   \\
        CIFAR+50      & 0.1, 0.05, 0.01   \\
        TinyImagenet   & 0.5, 0.1, 0.05 \\
    \bottomrule[1.5pt]
    \end{tabular}
    \vspace{5pt}
    \caption{Temperatures that are used in the final implementation.}
    \label{tab-Temperatures}
\end{table}

\subsection{CKA between Representations} \label{app-cka}

\begin{figure}[h]
    \centering
    \includegraphics[width=0.7\linewidth]{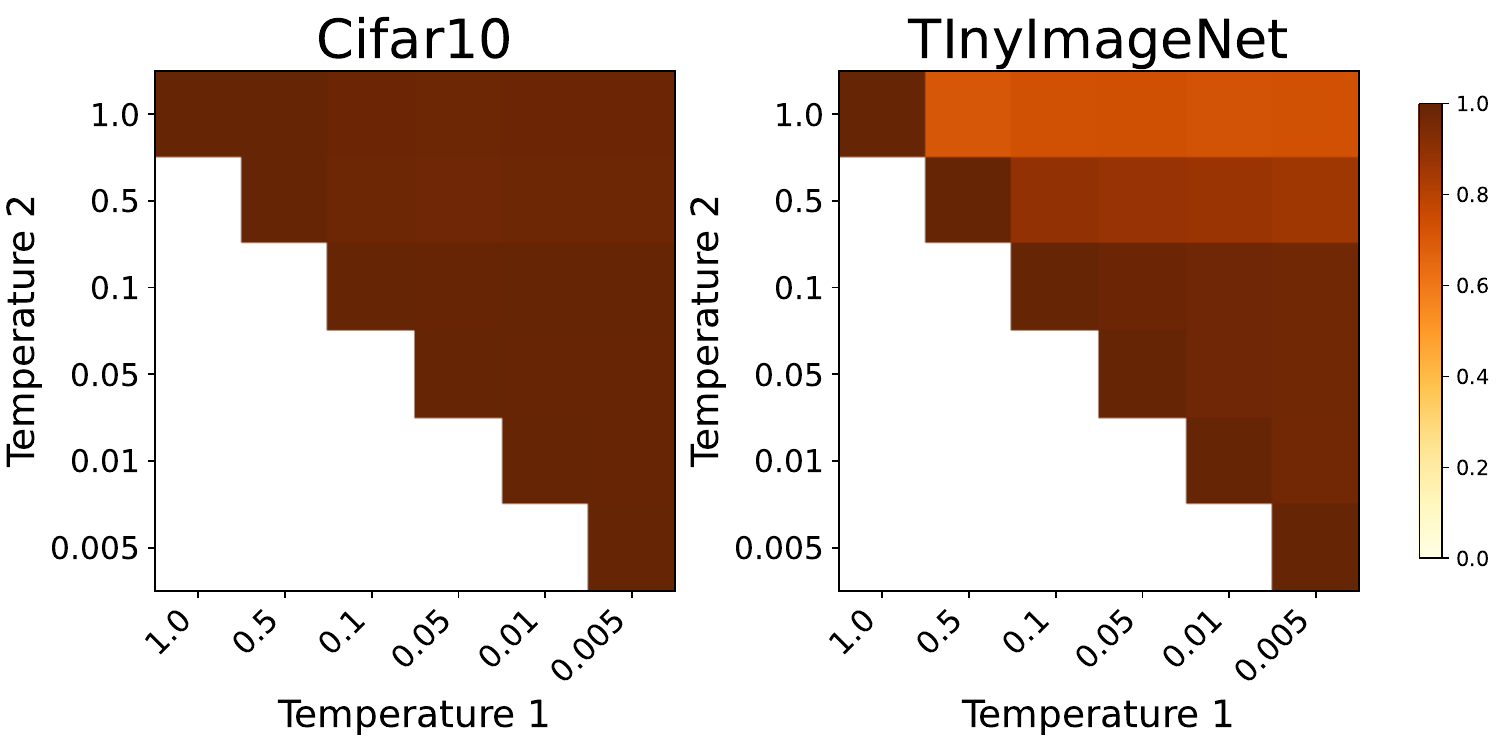}
    \vspace{3pt}
    \caption{CKA between the representations trained using variant temperatures. The average similarity is smaller for the TingImageNet dataset, indicating richer features to extract in the data.}
    \label{fig-cka}
\end{figure}

The centered kernel alignment (CKA) values between the representations with variant temperatures are demonstrated in Figure~\ref{fig-cka}. CKA measures the similarity between the representations. Smaller values mean less similarity. The similarities between the TingImageNet representations are overall smaller than those of CIFAR10, indicating their richer features in the data to learn.

\section{More on Ablations} \label{app-ab}

\begin{table}[h]
    \centering
\begin{tabular}{c|cc |c|cc }
    \toprule
         &	CIFAR10	& TinyImgNet &	&	CIFAR10	& TinyImgNet\\
         \hline
     1.0, 0.5    & $87.36$ &$78.42$ & 0.5, 0.005    & $92.79$ &$80.41$	 \\
     1.0, 0.1     & $87.77$ &$77.99$	& 0.1, 0.05    & $92.54$ &$79.33$   \\
     1.0, 0.05    & $87.86$ &$78.36$	& 0.1, 0.01    & $92.83$ &$78.66$ \\
     1.0, 0.01    & $87.91$ &$78.33$	& 0.1, 0.05    & $92.68$ &$79.24$ \\
     1.0, 0.005    & $88.01$ &$77.66$ & 0.1, 0.01 & $92.83$ &$78.66$ \\
     0.5, 0.1   & $91.84$ &$78.91$	& 0.1, 0.005    & $92.69$  &$77.86$ \\
     0.5, 0.05     & $92.82$ &$81.42$  & 0.05, 0.01    & $91.78$ &$78.01$\\
     0.5, 0.01     & $92.75$ &$79.36$   & 0.05, 0.005    & $91.45$ &$78.12$\\
     0.01, 0.005     & $91.34$ &$78.57$   &   &  &\\
     \bottomrule
    \end{tabular}
\vspace{5pt}
\caption{Complete results for two ensembles}
\label{tab-2}
    \end{table}

\begin{table}[h]
    \centering
    \scriptsize
    \begin{tabular}{c|c|c|c|c|c|c}
    \toprule
    & 1.0, 0.5, 0.1 &	1.0,0.5,0.05	& 1.0,0.5,0.01 & 1.0,0.5,0.005	&  1.0,0.05,0.01 & 1.0,0.05,0.005\\
         \hline
     CIFAR10    & $91.04$ &$91.56$	& $91.48$ & $91.35$  &	$91.57$ &  $93.47$   \\
     \hline
    TinyImgnet & $79.4$ & $79.66$	& $79.72$ & $79.64$ & $79.53$	&  $79.37$    \\
    \hline
         & 0.5,0.05,0.005 &	0.5,0.1,0.05	& 0.5,0.1,0.01 & 0.5,0.1,0.005	& 0.1,0.05,0.01 & 0.1,0.05,0.005   \\
         \hline
     CIFAR10    & $93.04$ &$92.96$	& $93.06$ & $92.9$  &	$\mathbf{93.12}$ &  $93$ \\
     \hline
    TinyImgnet & $80.3$ & $\mathbf{81}$	& $78.8$ & $79.48$ & $79.14$	&  $79.68$  \\
          \hline
      & 0.05,0.01,0.005 &		&  & 	& &    \\
         \hline
     CIFAR10    & $92.75$ &	&  &  &	 &  \\
     \hline
    TinyImgnet & $78.97$ & 	&  &  &  &    \\
     \bottomrule
    \end{tabular}
    \vspace{5pt}
    \caption{Complete results for three ensembles}
    \label{tab-3}
\end{table}

\hfill

\end{document}